\documentclass[conference]{IEEEtran}
\usepackage{amsmath,amsfonts}
\usepackage{algorithmic}
\usepackage{algorithm}
\usepackage{array}
\usepackage[caption=false,font=normalsize,labelfont=sf,textfont=sf]{subfig}
\usepackage{textcomp}
\usepackage{stfloats}
\usepackage{url}
\usepackage{verbatim}
\usepackage{graphicx}
\usepackage{cite}
\usepackage{cuted} 
\usepackage{tabularx}
\usepackage{adjustbox}
\usepackage{booktabs}
\usepackage{multirow}
\usepackage{threeparttable}
\usepackage{pifont}
\usepackage{tabularx}
\IEEEoverridecommandlockouts   

\usepackage{xcolor,soul,framed} 
\colorlet{shadecolor}{yellow}

\begin{document}

\title{DTCCL: Disengagement-Triggered Contrastive Continual Learning for Autonomous Bus Planners}

\author{
  Yanding Yang,
  Weitao Zhou,
  Jinhai Wang,
  Xiaomin Guo,
  Junze Wen,
  Xiaolong Liu,
  Lang Ding,\\
  Zheng Fu,
  Jinyu Miao,
  Kun Jiang,
  and Diange Yang

\thanks{Yanding~Yang,~Weitao~Zhou,~Zheng~Fu,~Jinyu~Miao,~Junze~Wen,~Kun~Jiang, and Diange~Yang are with the School of Vehicle and Mobility, 
Tsinghua University, Beijing~100084,~China 
}
\thanks{Yanding~Yang and Jinhai~Wang, are with Dongfeng Motor Corporation Research and Development General Institute, Wuhan~430056,~China 
}

\thanks{Xiaomin~Guo,~Xiaolong~Liu,~and~Lang~Ding are with Dongfeng~Yuexiang~Technology~Co.,~Ltd., Wuhan~430000,~China 
}
\thanks{Corresponding authors: Diange~Yang (ydg@mail.tsinghua.edu.cn) and Weitao~Zhou (zhouwt@tsinghua.edu.cn).}
}

\maketitle

\begin{abstract}
Autonomous buses run on fixed routes but must operate in open, dynamic urban environments. Disengagement events on these routes are often geographically concentrated and typically arise from planner's failures in highly interactive regions. Such policy-level failures are difficult to correct using conventional imitation learning, which easily overfits to sparse disengagement data. To this end, this paper presents a Disengagement-Triggered Contrastive Continual Learning (DTCCL) framework that enables autonomous buses to improve driving policies of planner through real-world operation. Each disengagement triggers cloud-based data augmentation that generates positive and negative samples by perturbing surrounding agents while preserving route context. Contrastive learning refines policy representations to better distinguish safe and unsafe behaviors, and continual updates are applied in a cloud–edge loop without human supervision. 
Experiments on urban bus routes demonstrate that DTCCL improves overall planning performance by 48.6\% compared with direct retraining, validating its effectiveness for scalable, closed-loop policy improvement in autonomous public transport.

\end{abstract}


\section{Introduction}

In recent years, assisted driving systems have achieved remarkable progress in passenger vehicles, demonstrating reliable performance under diverse urban conditions.  This success has inspired expectations that fully driverless operation could soon be realized in more structured settings—such as autonomous buses operating along fixed routes. However, the reality remains more complex. Although these vehicles follow predefined routes and geofenced corridors, they must still navigate open and dynamically changing urban environments characterized by dense traffic, unpredictable pedestrians, and complex multi-agent interactions. Owing to the fixed routes, both high-definition maps and perception modules can be carefully optimized in advance, thus many disengagements observed in real-world operations originate from the planning layer. These planner-level failures are often concentrated at specific geographic locations, such as unprotected left turns or busy merging points, where the system must balance safety with operational efficiency. Leveraging these disengagement data to refine the driving policy of planner remains a core challenge for large-scale deployment of autonomous bus fleets. 

Most existing planning policies for autonomous driving are trained offline through imitation learning, assuming that a fixed dataset of expert demonstrations is sufficient to generalize across diverse environments~\cite{codevilla2018end, bojarski2016end}. This paradigm performs well in closed or richly annotated settings but lacks adaptability to new cases of real-world deployment. Autonomous buses, despite operating on fixed routes, must handle open, interactive urban environments. In these settings, perception and map priors can be optimized offline, yet most disengagements during operation stem from planner-level failures at recurrent geographic hotspots such as unprotected turns or narrow merging lanes.
These disengagement events represent natural feedback from real-world operation. However, conventional offline learning cannot exploit such feedback, and directly fine-tuning on sparse disengagement data often leads to overfitting and unstable updates~\cite{cao2022autonomous, zhou2025drarl}. Bridging this gap requires a closed-loop learning paradigm that continuously integrates disengagement-triggered supervision into policy refinement, which remains an open challenge for driving policy improvement \cite{cao2023continuous, mei2024continuously}.

\begin{figure}
    \centering
    \includegraphics[width=1\linewidth]{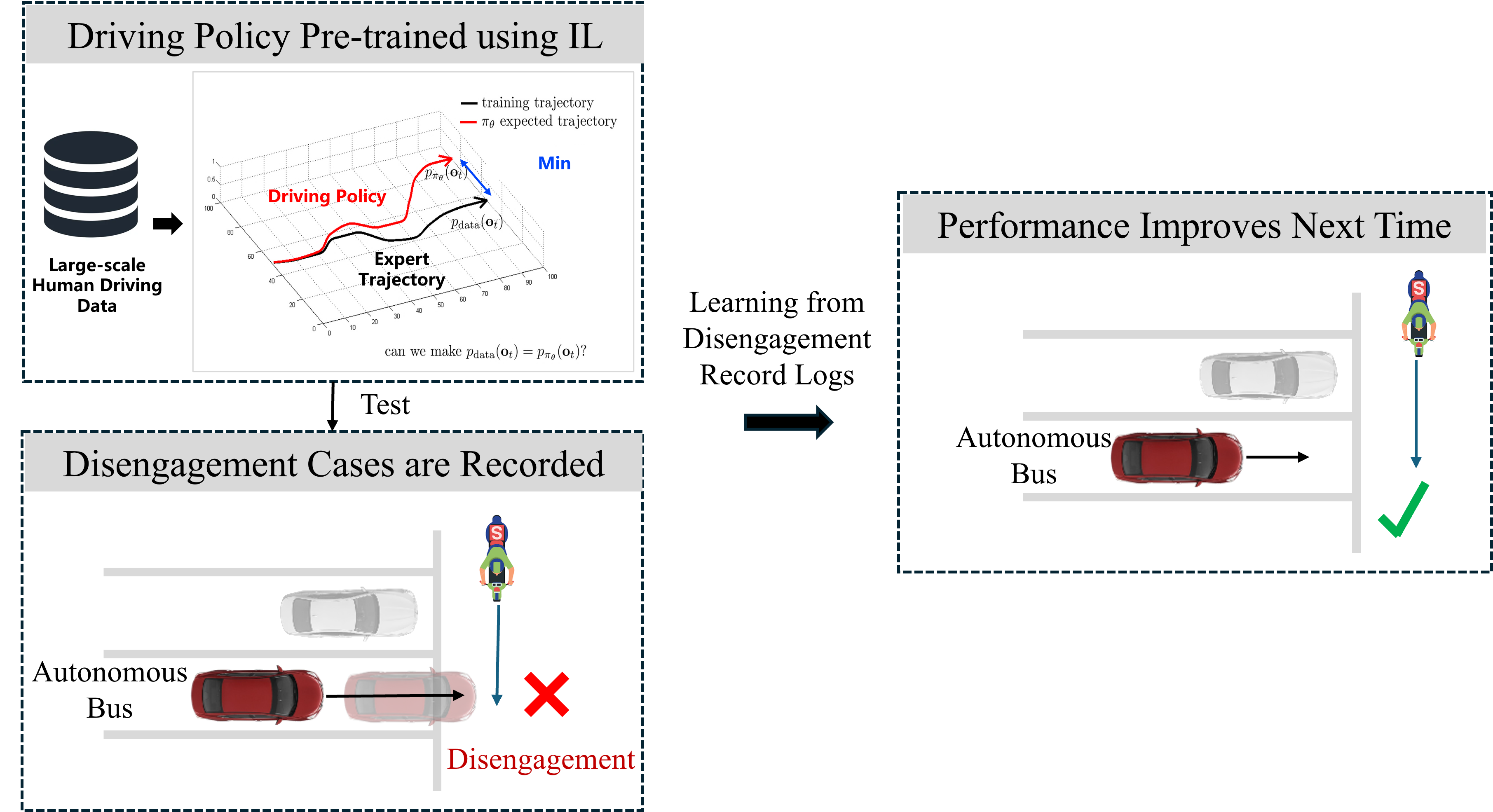}
    \caption{Closed-loop continual learning framework for autonomous bus planning.  The driving policy is first pretrained via imitation learning on large-scale human driving data. During deployment, disengagement events are automatically recorded and used to trigger policy learning, enabling the planner to achieve improved performance when encountering similar scenarios in the future.  }
    \label{fig:idea}
\end{figure}

To this end, we propose a fully automated policy improvement framework, which is structured as a ``Collection–Augmentation–Regression–Validation'' pipeline. It begins by detecting disengagement events triggered during deployment on real-world bus lines, capturing the scenario logs before and after human intervention. These data slices are then used to reconstruct rare corner-case scenarios through contrastive learning and safety-aware augmentation. Then the driving policy is iteratively updated to adapt to similar future cases. The learning process is deployed in the cloud and operates autonomously across an entire bus fleet, minimizing the need for manual labeling or engineering effort. 

Our contributions are summarized as follows:

1) Disengagement-driven continual learning framework:
We introduce a continual learning framework that progressively enhances a pretrained imitation learning planner’s performance in recurring real-world conditions. It transforms planner-level failures in autonomous bus fleets into explicit supervision signals, establishing a closed-loop process for feedback-based policy improvement.

2) Contrastive data augmentation for disengagement logs:
We propose a data augmentation method that reconstructs semantically consistent positive and negative samples from disengagement cases by perturbing surrounding agents while maintaining route context, enhancing robustness in sparse and safety-critical scenarios.

3) Contrastive continual learning method for autonomous bus planner:
We develop a contrastive continual learning approach that fine-tunes the pretrained policy using augmented disengagement data. The method jointly optimizes imitation and contrastive objectives to refine planner representations while preserving previously learned behaviors, achieving stable and progressive improvement in planner performance under real-world operational feedback.



\section{Related Works}


\subsection{Learning-based Planner of Autonomous Bus}

The development of planning and control for autonomous buses combines classical optimization with learning-based planners to handle unique challenges such as passenger comfort, frequent stops, and dense urban interactions\cite{liu2021decision}. 

While model-based methods have been widely used for trajectory planning and control, imitation learning (IL) has become a dominant paradigm for data-driven policy generation, enabling direct learning from expert demonstrations\cite{kuderer2015learning, codevilla2018end, bojarski2016end, chen2015deepdriving}. 
Recent IL frameworks such as PLUTO\cite{cheng2024pluto} and PlanTF\cite{cheng2024rethinking} further enhance representation stability and training robustness through large-scale datasets and transformer-based architectures, offering strong priors for urban driving policies. 

In contrast, reinforcement learning (RL) approaches improve adaptability through environment interactions\cite{sallab2017deep, nie2024robust}, allowing planners to optimize both safety and efficiency in complex traffic. 
Adaptive or conservative RL planners\cite{zhou2022dynamically} can adjust decision aggressiveness in long-tail cases, while hierarchical RL methods\cite{gu2023safe, naveed2021trajectory} enhance safety and stability in intersection and merging tasks. 
However, most existing IL- and RL-based planners rely on static datasets or simulation-only training, limiting their ability to adapt to real-world feedback, especially in fixed-route autonomous bus deployments where disengagement data is sparse but safety-critical.

\subsection{Continual Learning in Autonomous Driving Planner}

Continual learning (CL) enables incremental policy refinement under evolving environments after the policy is initialized or pre-trained\cite{parisi2019continual, rusu2016progressive}. 
Replay-based methods maintain buffers of past experience to stabilize learning\cite{wu2022prioritized, smith2024adaptive}, while regularization-based strategies constrain weight updates to preserve prior knowledge\cite{kang2024continual}. 
Contrastive learning has recently been integrated into continual frameworks to improve latent representation robustness\cite{zhang3cil, ma2022compass, xiao2021action}, helping policies distinguish subtle but safety-critical scene variations. 

In autonomous driving, CL has been explored for continuous adaptation using simulation-driven or imagination-based replay\cite{cao2022autonomous, mei2024continuously}, yet validation largely remains within passenger-vehicle contexts. Studies analyzing real-world disengagement logs\cite{favaro2018autonomous, dixit2016autonomous} reveal that operational intervention data carries critical cues for improving planning reliability. 
Nevertheless, few works have systematically leveraged disengagement events as direct feedback signals for planner improvement, particularly in structured yet dynamic scenarios like bus fleets.

\subsection{Data Augmentation for Few-shot Learning}

Data augmentation is an effective means to improve model generalization under limited or imbalanced data conditions. 
Domain randomization (DR) and simulation-based augmentation\cite{hu2023simulation, niu2021dr2l, mehta2020active, zuo2025ralad} generate diverse environments by varying lighting, weather, or traffic configurations, while active DR\cite{DAOUD2024184} and meta-learning\cite{yu2024using} further enhance adaptation to unseen cases. 
Recently, large language model (LLM)-driven frameworks\cite{10529537} have been proposed to synthesize complex driving scenes and rare event distributions. 
However, simulation- or LLM-generated data often diverge from the semantics of real operational corner cases. 
In contrast, this work performs safety-aware data augmentation directly on real disengagement logs, reconstructing semantically consistent positive and negative samples through agent perturbations within the same route context, thereby capturing true long-tail distributions faced by autonomous buses.

\section{Preliminaries and Problem Setup}

\subsection{Preliminaries}

A widely adopted approach to initializing autonomous driving policies is imitation learning, where a policy $\pi_\theta$ is trained to imitate expert demonstrations via supervised learning. Given a dataset of state–action pairs $\{(s_t, a_t)\}$ collected from human drivers, the policy aims to minimize the following imitation loss:

\begin{equation}
\mathcal{L}_{\mathrm{BC}} = \mathbb{E}_{(s, a) \sim \mathcal{D}_{\text{human}}} \left[ \left\| \pi_\theta(s) - a \right\|^2 \right]
\end{equation}

This pretraining process enables the policy to acquire general driving skills across nominal scenarios. When deployed in the real world, however, the policy inevitably encounters  performance degradation and policy failures in long-tail cases. Such failures are typically manifested as \textit{disengagements}—moments when a human driver intervenes to override the system. 
Each disengagement provides a high-value sample $(s_t^d, a_t^d)$, representing the deviation between expert and policy behavior:
\begin{equation}
\mathcal{D}_{\text{dis}} = 
\{(s_t^d, a_t^d) \mid 
a_t^d \neq \pi_\theta(s_t^d),\,
\text{disengagement at } t \}.
\end{equation}


\subsection{Problem Setup}

Given a pretrained policy $\pi_0$ obtained from large-scale human driving data via imitation learning, and a set of disengagement segments $\{\mathcal{S}_i\}$ collected from real-world autonomous bus deployments, the goal is to derive an improved policy $\pi^*$ that:
\begin{itemize}
    \item Reduces the likelihood of future disengagements in similar contexts;
    \item Learns from limited high-value feedback to generalize to unseen but related failure modes.
\end{itemize}

Each disengagement segment $\mathcal{S}_i$ captures the temporal evolution of system behavior around a disengagement event, providing both contextual information and corrective supervision. These samples serve as the foundation for our subsequent data augmentation and contrastive continual learning framework.

\section{Method}

This section introduces the proposed closed-loop planning framework for autonomous buses. The framework leverages large-scale imitation learning for pretraining, real-world disengagement mining for discovering failure cases, positive/negative sample generation for data augmentation, and contrastive learning for reliable driving policy learning. Finally, the adapted policy is optimized with a joint objective to progressively improve performance in rare and challenging scenarios.

\subsection{Overview of the Framework}
\begin{figure}[t]
    \includegraphics[width=3.3in]{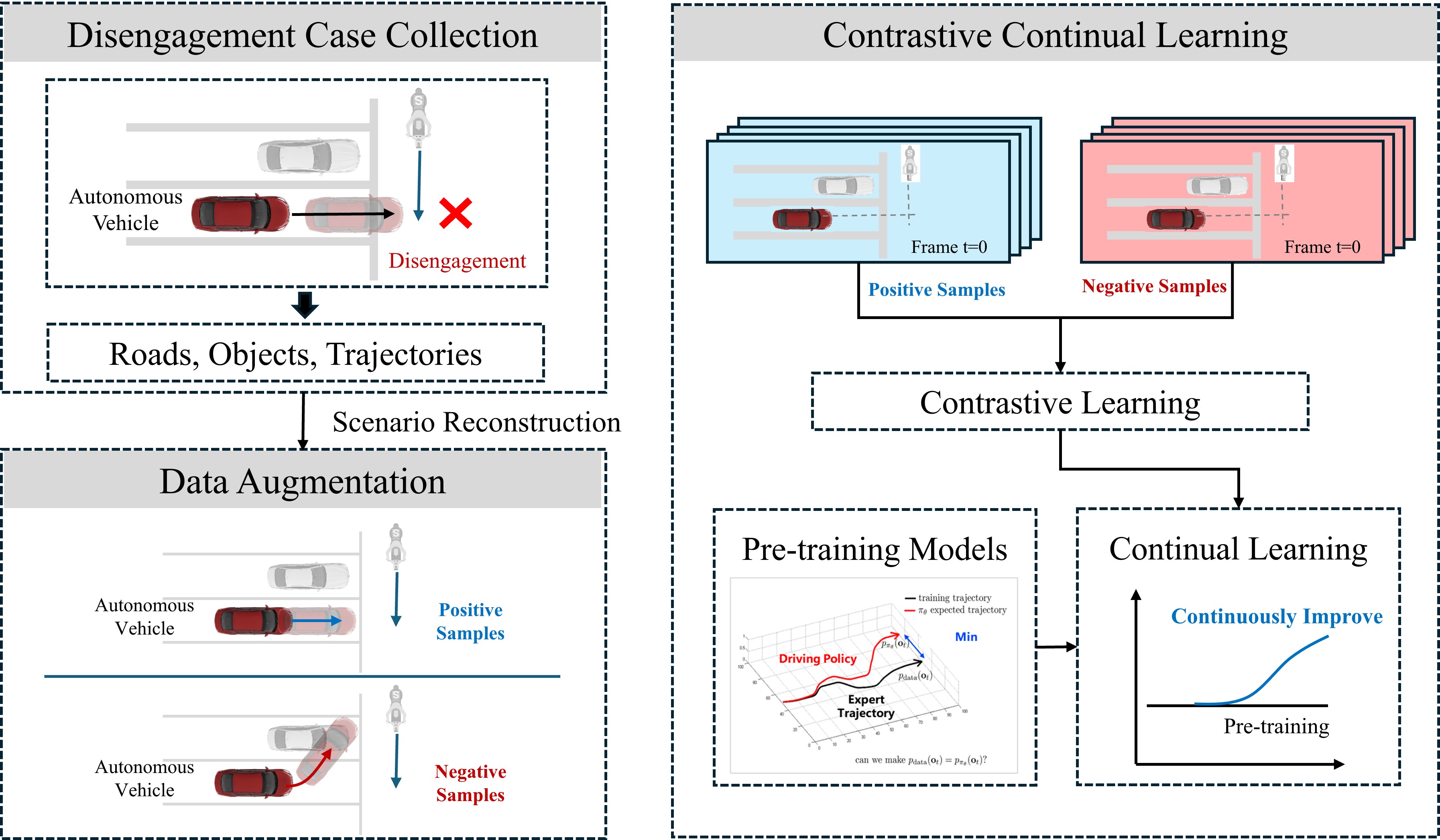}
    \caption{The framework of proposed DTCCL method. A base policy is first pretrained via imitation learning on large-scale human driving data. During deployment, disengagement events are recorded and transformed into positive and negative variants through safety-aware augmentation. Contrastive learning is then applied to improve representation robustness, enabling continual policy refinement across deployment cycles.}
    \label{framework}
\end{figure}

As illustrated in Fig.~\ref{framework}, our framework follows a ``Collect–Augment–Adapt'' pipeline. First, a base policy $\pi_0$ is pretrained using imitation learning on large-scale human driving data, capturing general driving behavior across typical conditions. During deployment, the autonomous bus fleet inevitably encounters disengagements, where a safety driver intervenes. We record each disengagement and apply safety-aware augmentation to generate positive and negative variants that retain semantic and contextual consistency. Subsequently, a contrastive learning objective is used to enhance the discriminability of policy embeddings under these augmented scenarios. Finally, the policy continually improves over deployment cycles, resulting in safer and more reliable autonomous bus operation.

\subsection{Disengagement-Triggered Data Augmentation}

\subsubsection{Disengagement Mining}
During deployment, the system continuously monitors the operational state of the autonomous bus to detect transitions between autonomous driving and human takeover. A disengagement event is defined as a consistent switch from a stable autonomous driving phase to sustained manual intervention, ensuring the detection corresponds to a genuine policy-level failure rather than transient noise. Once such a transition is identified, a short temporal segment centered around the intervention time $t_d$ is extracted:
\begin{equation}
\mathcal{S} = \{(s_{t-k}, a_{t-k}), \ldots, (s_t, a_t), \ldots, (s_{t+k}, a_{t+k})\}.
\end{equation}

The segment captures both the preceding and subsequent contexts surrounding the disengagement. We particularly retain the frames before $t_d$, as planner-induced failures typically emerge prior to human disengagements, whereas states at or after $t_d$ may be contaminated by emergency manual maneuvers. These extracted temporal slices serve as high-value supervision anchors for subsequent data augmentation and contrastive continual learning.

\paragraph{Positive Sample Generation }
Within the pre-disengagement window $\,[\tau_a,\,t_d)$, we construct positive variants that remain feasible under the safety checker and preserve the fixed-route semantics through two strategies.

(1) Safety-constrained ego micro-perturbations:
Small perturbations are applied to the ego state \(\delta x=(\delta p_x,\delta p_y,\delta \psi,\delta v)\) starting from $\tau_a$, subject to backoff–retry to ensure: (a) collision-free interaction with nearby agents over a short horizon, (b) full compliance with drivable-area constraints, and (c) adherence to dynamic limits on yaw rate, acceleration, and jerk. These controlled perturbations generate recoverable trajectories near the decision boundary, allowing the policy to learn smooth corrections before failure occurs.

(2) Non-interactive agent dropout with semantic preservation:
To reduce overfitting to background clutter, agents exerting negligible influence on the ego decision are stochastically removed while preserving lead and priority vehicles. Interactive agents are identified by the overlap of their predicted future bounding boxes with the ego trajectory within a fixed temporal horizon. For dropped agents, scene integrity is maintained by preserving lane topology, traffic rules, and timing consistency of the remaining participants, followed by temporal smoothing to avoid discontinuities. This operation filters out irrelevant context while retaining realistic scene semantics, focusing the model’s attention on genuinely interactive cues.

Together, these two strategies yield semantically consistent, safety-compliant variants of disengagement segments, providing informative positive supervision around previously failure-prone contexts.

\paragraph{Negative Sample Generation ($D_{\text{neg}}$).}
To explore the unsafe side of the decision boundary, we deliberately amplify failure-inducing factors within the pre-disengagement window. This process constructs synthetic yet physically plausible trajectories that expose the policy to diverse unsafe behaviors:  
(1) lead-agent insertion, where a slow or braking vehicle is introduced along the ego path to induce headway violations and test reaction capacity;  
(2) lead-agent removal, which eliminates critical forward vehicles to examine whether the policy can still identify unsafe gaps without explicit cues;  
(3) interactive-agent dropout, removing highly interactive agents to produce unrealistic dynamics that probe the model’s stability and generalization limits; 
(4) large-magnitude ego perturbations, directly pushing the ego trajectory into collision or violation zones according to the safety checker.  
These operations jointly expand the coverage of high-risk conditions, enabling the policy to explicitly distinguish between recoverable and failure regimes.

The final augmented dataset is defined as:
\[
D_{\text{aug}} = D_{\text{human}} \cup D_{\text{pos}} \cup D_{\text{neg}},
\]
where $D_{\text{human}}$ denotes the large-scale human driving corpus used for pretraining.

\subsection{Contrastive Learning for Disengagement Cases}

\subsubsection{Representation Learning with Contrastive Objectives}

To refine the pretrained policy $\pi_0$ using disengagement-triggered data, we employ a contrastive representation learning strategy that structures the latent space around safety-critical semantics. The objective is to align safe augmentations with their anchor states while separating unsafe ones, thereby enhancing policy robustness under disengagement cases.

Given a disengagement-centered sample $x$, we generate a positive variant $x^{+}$ and a negative variant $x^{-}$ using data augmentation operators described in last section. All three samples are processed through a shared Transformer encoder $h(\cdot)$ and a two-layer projection head $g(\cdot)$ to obtain latent embeddings:
\begin{equation}
z = g(h(x)), \quad z^{+} = g(h(x^{+})), \quad z^{-} = g(h(x^{-})).
\end{equation}

A triplet contrastive loss is computed to enforce similarity between the anchor $z$ and the positive $z^+$, while decreasing the similarity to the negative $z^-$:
\begin{equation}
\mathcal{L}_{CL} = - \log \frac{\exp(\text{sim}(z, z^+)/\tau)}{\exp(\text{sim}(z, z^+)/\tau) + \exp(\text{sim}(z, z^-)/\tau)},
\end{equation}
where $\text{sim}(u,v) = \tfrac{u^\top v}{\|u\|\|v\|}$ is the cosine similarity and $\tau$ is a temperature parameter. This loss is computed across all triplets in a minibatch, making the optimization effective even under limited disengagement data.

In practice, each batch of $N_{bs}$ samples undergoes both $\mathcal{T}^{+}$ and $\mathcal{T}^{-}$ transformations, yielding $3N_{bs}$ instances per iteration. The shared encoder ensures consistent feature extraction, while the projection head provides flexibility for continual adaptation. Importantly, negatively augmented samples are used solely for contrastive learning, since their trajectories lose causal validity after augmentation. Conversely, the original and positively augmented samples remain supervised by human corrective actions through imitation learning. This dual objective couples representation robustness with behavioral fidelity, enabling the refined policy to distinguish recoverable from unsafe states while imitating expert recovery strategies.

\subsection{Driving Policy Continual Improvement Process}
The proposed framework operates as a closed-loop continual improvement system, where the policy is periodically refined based on disengagement data collected from real-world operations. This process runs automatically in a cloud–edge loop, following the sequence: \textit{Deploy} → \textit{Collect} → \textit{Augment} → \textit{Update} → \textit{Redeploy}.

During deployment, the current policy $\pi_t$ is executed on autonomous buses. Whenever a disengagement occurs, surrounding sensor data, ego states, and corrective human actions within a fixed temporal window are logged as a scenario batch $\mathcal{B}_t$. 

These batches are uploaded to the cloud for processing once sufficient events accumulate. In the cloud, $\mathcal{B}_t$ is enhanced via the proposed disengagement-triggered augmentation strategy to produce $\mathcal{D}_{\text{aug}}^{(t)}$. The updated policy $\pi_{t+1}$ is then obtained by fine-tuning $\pi_t$ with the objective:
\begin{equation}
\pi_{t+1} = \arg\min_{\pi} \big[w_1 \mathcal{L}_{BC}^{dis} + w_2 \mathcal{L}_{CL} + w_3 \mathcal{L}_{stab}\big],
\end{equation}
where $\mathcal{L}_{BC}^{dis}$ uses human corrective actions for imitation, $\mathcal{L}_{CL}$ ensures contrastive consistency, and $\mathcal{L}_{stab}$ constrains deviation from $\pi_t$ to maintain stability. After validation on nominal and disengagement subsets, $\pi_{t+1}$ is redeployed to resume data collection, forming a continual improvement cycle.

\begin{algorithm}[t]
\caption{Driving Policy Continual Improvement Loop}
\label{alg:continual}
\begin{algorithmic}[1]
\STATE Initialize pretrained policy $\pi_0$.
\FOR{$t=0,1,2,\dots$}
  \STATE Deploy $\pi_t$ on the autonomous bus fleet.
  \STATE Collect disengagement scenarios $\mathcal{B}_t$.
  \STATE Augment $\mathcal{B}_t$ to form $\mathcal{D}_{\text{aug}}^{(t)}$ using positive/negative generation.
  \STATE Obtain $\pi_{t+1}$ by minimizing $\mathcal{L}=w_1\,\mathcal{L}_{BC}^{dis}+w_2\,\mathcal{L}_{CL}+w_3\,\mathcal{L}_{stab}$.
  \STATE Validate $\pi_{t+1}$ on safety gates (nominal \& disengagement subsets).
  \IF{validation passed}
    \STATE Redeploy $\pi_{t+1}$ and continue.
  \ELSE
    \STATE Retain $\pi_t$ and keep logging for the next batch.
  \ENDIF
\ENDFOR
\end{algorithmic}
\end{algorithm}

This process allows the policy to self-improve through real operational data, gradually reducing disengagement frequency and enhancing robustness in long-tail scenarios without manual supervision.


\section{Case Study}

This section presents a comprehensive evaluation of the proposed disengagement-driven policy continual learning framework for autonomous bus planner. We first describe the experimental setup, including the simulation environment, pretraining dataset, and the procedure for collecting real-world disengagement data. Next, we outline the evaluation metrics designed to measure improvements in safety, stability, and robustness under long-tail scenarios. The main experimental results compare the proposed approach with several baseline methods, quantifying the benefits of disengagement-guided learning in both simulated and real-world settings. Finally, case studies illustrates the system’s behavior in complex urban traffic scenarios.

\subsection{Experiment Setup}

\subsubsection{Autonomous Bus Platform and Driving Policy Pre-training}

\begin{figure}[h]
    \includegraphics[width=\linewidth]{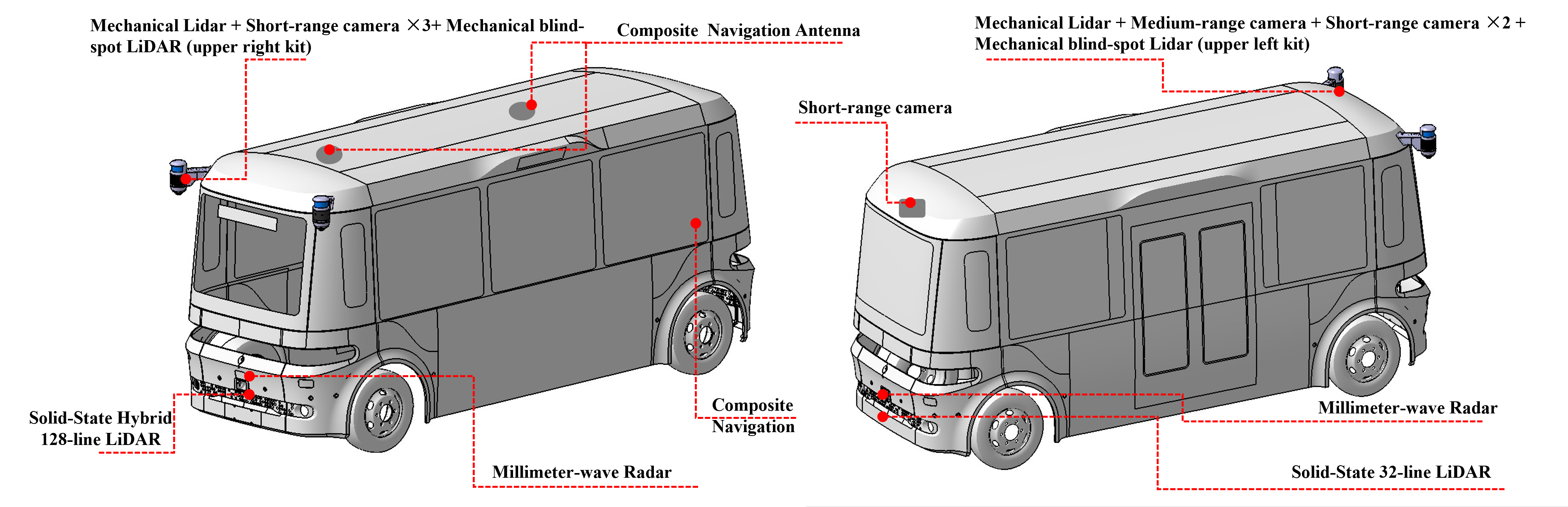}
    \caption{Autonomous bus platform equipped with multi-sensor perception, including LiDARs, short- and medium-range cameras, millimeter-wave radars, and composite navigation modules.}
    \captionsetup{justification=centering}
    \label{autonomous}
\end{figure}

All experiments are conducted on a full-scale autonomous bus platform equipped with a multi-sensor perception and navigation system, as shown in Fig~\ref{autonomous}. The vehicle integrates mechanical and solid-state LiDARs, millimeter-wave radars, and short- and medium-range cameras to provide 360° perception coverage. Dual roof-mounted sensor modules enable redundancy for environment sensing, while front and rear solid-state LiDARs enhance detection accuracy at close range. A composite navigation unit combining RTK-GNSS and IMU ensures centimeter-level localization accuracy, providing a reliable foundation for data collection and policy evaluation in real-world operations.

The driving policy of autonomous bus is initialized using the PLUTO framework~\cite{cheng2024pluto} and pretrained on the large-scale nuPlan dataset~\cite{karnchanachari2024towards}. The training dataset contains approximately one million urban driving samples covering diverse and complex scenarios. This stage establishes a robust baseline policy capable of handling nominal urban conditions\cite{qian2024spider}.

\subsubsection{Disengagement Cases Extraction}

To enable real-world adaptation, autonomous buses equipped with the full sensor suite are deployed along fixed routes, where disengagement events are automatically logged for subsequent policy refinement. The onboard system monitors frame-level control status labeled as \texttt{cruise} (autonomous) or \texttt{off} (manual override). 
A original disengagement case is triggered when 30 consecutive \texttt{cruise} frames are immediately followed by 20 consecutive \texttt{off} frames, forming a 50-frame temporal segment centered on the final \texttt{off} frame ($t$), which is defined as the moment of takeover. This window captures both pre- and post-intervention contexts for downstream augmentation and learning.

In real-world autonomous bus operations, not all disengagements correspond to planner failures—many are casual interventions unrelated to policy quality. To isolate cases that genuinely reflect low policy performance, all collected disengagement events were reconstructed and replayed in an offline simulation environment using the pretrained policy. This replay-based verification allowed us to identify scenarios where the planner consistently failed, resulting in a subset of 756 representative failure cases. These selected cases form the training set for continual learning.

\subsubsection{Cloud-based Model Update and Evaluation}

All model training and updates are conducted on a cloud server equipped with several NVIDIA A800 GPUs (80 GB memory each). After deployment, the autonomous bus fleet continuously uploads disengagement data—including sensor recordings, ego states, and human corrective actions—to the cloud once sufficient samples have been accumulated. 

The cloud server executes the full model update pipeline, consisting of disengagement-triggered data augmentation, contrastive representation learning, and continual fine-tuning based on the updated dataset $\mathcal{D}_{\text{aug}}$. 

Each updated policy $\pi_{t+1}$ is validated on  disengagement-centered test sets containing over 2,500 real-world scenarios. Each scenario corresponds to a past disengagement event recorded during autonomous bus operations, representing cases where human intervention was required. 
These scenarios include dense pedestrian crossings, unprotected left turns, narrow lanes, and other high-interaction regions typical of urban traffic. 
By emphasizing safety-critical disengagement test cases, this dataset provides a realistic benchmark for assessing continual learning ability after learning from disengagement records.

\subsection{Baselines}

To benchmark the proposed DTCCL framework, we compare it against three  baselines that differ in continual learning methods, that is, their use of disengagement data, data augmentation, and contrastive learning. The configurations of each strategy are summarized in Table~\ref{tab:method_comparison}. All models are trained on two NVIDIA A800 GPUs (80~GB each) for 25 epochs with a batch size of 128.

\begin{table}[t]
\centering
\caption{Comparison of Policy Learning Baselines}
\label{tab:method_comparison}
\resizebox{\columnwidth}{!}{%
\begin{tabular}{@{}lccc@{}}
\toprule
\textbf{Method} & \textbf{Disengagement Data} & \textbf{Data Aug.} & \textbf{Contrastive Learning} \\ \midrule
Direct IL & \ding{51} & \ding{55} & \ding{55} \\
Augmented IL & \ding{51} & \ding{51} & \ding{55} \\
\textbf{DTCCL (ours)} & \ding{51} & \ding{51} & \ding{51} \\ \bottomrule
\end{tabular}%
}
\end{table}

\paragraph{Direct IL.}
This baseline fine-tunes the Pluto policy directly on the real disengagement dataset without any augmentation or contrastive objective. It reflects the impact of directly leveraging raw disengagement data for adaptation.

\paragraph{Augmented IL.}
In this variant, the policy is fine-tuned on disengagement data combined with positive data augmentations including non-interactive agent dropout and ego perturbations. These augmentations are used to train the driving policy with imitation learning methods.

\paragraph{DTCCL (ours).}
The proposed method extends the above setup by introducing disengagement-triggered semantic augmentation and contrastive continual learning. 

Together, these baselines provide a progressive comparison spectrum—from direct imitation fine-tuning to full contrastive continual learning—allowing a quantitative assessment of each component’s contribution to the overall policy improvement.

\subsection{Performance Metrics}

Following the evaluation design of the nuPlan benchmark suite~\cite{caesar2021nuplan}, 
we adopt a set of closed-loop metrics emphasizing safety, comfort, and driving efficiency. Each episode of disengagement cases is evaluated using nuPlan-style indicators, including \emph{Collisions}, \emph{Drivable area compliance}, \emph{Direction adherence}, \emph{Time-to-Collision (TTC)}, \emph{Comfort}, \emph{Speed limit compliance}, 
and \emph{Progress} along the planned route. 

All metrics are computed in our in-house closed-loop simulation platform that reproduces realistic bus-scale dynamics and traffic conditions. 
A composite score, consistent with nuPlan’s weighting scheme, is used to summarize overall performance. This provides a unified indicator for quantitative comparison across different models and facilitates consistent benchmarking of policy improvement.

\subsection{Main Results and Discussion}



\begin{table*}[t]
\caption{Evaluation Results of Different Driving Policies in Testing Scenarios}
\label{tab:results}
\centering
\footnotesize
\resizebox{\linewidth}{!}{
\begin{tabular}{@{}lcccccccc@{}}
\toprule
Driving Policy &
  Score &
  Collisions (\%) &
  Drivable (\%) &
  Direction (\%) &
  TTC &
  Comfort &
  Speed &
  Progress \\ 
\midrule
Pre-trained IL         & 44.56 & 94.61 & 98.64 & 100 & 42.87 & \textbf{61.28}& 99.99 & 41.08 \\
Direct IL    & 51.53 & 94.49 & \textbf{99.16}& 100 & 60.04& 60.88& 99.39& 44.50\\
Augmented IL     & 51.23 & 94.05 & 92.22 & 100 & 60.72& 61.00& 98.99& 37.74\\
Ours (DTCCL)         & \textbf{69.66} & \textbf{96.48} & 99.02& \textbf{100} & \textbf{85.67} & 58.70 & \textbf{100.00} & \textbf{75.42} \\
\bottomrule
\end{tabular}}
\vspace{1.2mm}
\footnotesize
\end{table*}

We evaluate all methods on the disengagement-focused dataset introduced above, which emphasizes real-world disengagement cases in autonomous bus operations. The results of this evaluation are presented in TABLE~\ref{tab:results}. 

Across all metrics, the proposed method demonstrates consistent improvements in both safety and operational efficiency. 
Compared with the pre-trained imitation learning (IL) baseline, the proposed policy achieves a 48.6\% increase in the overall performance score, along with substantial gains in Time-to-Collision, Speed, and Progress metrics. 
These improvements reflect the enhanced planning performance through contrastive continual learning.

Although the comfort score is slightly lower than that of the baselines, this trade-off represents a more conservative and proactive driving strategy, which is desirable in public transportation where safety margins are prioritized over passenger comfort. 
The DTCCL policy also achieves the lowest collision rate among all evaluated planners, confirming that disengagement-guided learning effectively reduces risks in failure-prone conditions.

The Data-Augmented IL Planner attains the second-best performance, validating the benefit of disengagement-based scenario reconstruction even without explicit contrastive learning. 
However, its limited ability to differentiate recoverable from non-recoverable states leads to less stable decision-making under uncertain conditions. 
In contrast, DTCCL explicitly structures policy representations around safety semantics, yielding more robust and adaptive behavior in long-tail and highly interactive urban scenarios.

Overall, these results confirm that transforming disengagement data into structured supervision enables autonomous buses to continuously refine their planning policies. 
The proposed closed-loop learning paradigm provides measurable gains in safety and efficiency, demonstrating its potential for scalable deployment in real-world autonomous bus operations.

\subsection{Case Study}

To further illustrate how disengagement-driven continual learning improves policy behavior in safety-critical situations, we present two representative real-world cases extracted from the validated dataset. 

\paragraph{\textbf{Case 1 — Red-light misjudgment at an intersection.}}

The previous driving policy incorrectly generated a forward-driving trajectory despite the perception module having correctly detected a red traffic light. This led to a disengagement to prevent a potential violation. After DTCCL training, the refined policy produces a clear stopping decision in the same scenario, aligning its trajectory with the perceived signal state and avoiding risky acceleration toward the intersection. This shows the model’s improved ability to bind high-level semantic cues (traffic signals) with safe motion plans. 

\paragraph{\textbf{Case 2 — Excessive lateral avoidance near roadside guardrails.}}

In this case, the initial policy executed an overly conservative left-avoidance maneuver when overtaking a slow-moving leading vehicle, causing the bus to drift dangerously close to a roadside guardrail. This triggered a disengagement due to the risk of scraping or collision. After learning from disengagement-centered augmented samples, the updated policy achieves a more balanced lateral response: it successfully avoids the leading vehicle while maintaining a safe clearance from the guardrail. This demonstrates enhanced interaction handling in constrained road geometry.

Across both cases, the learned policy exhibits more accurate decision boundaries in long-tail, safety-critical environments, validating that disengagement-triggered contrastive continual learning leads to practical improvements in real-world autonomous bus planning.

\begin{figure}
    \centering
    \includegraphics[width=1\linewidth]{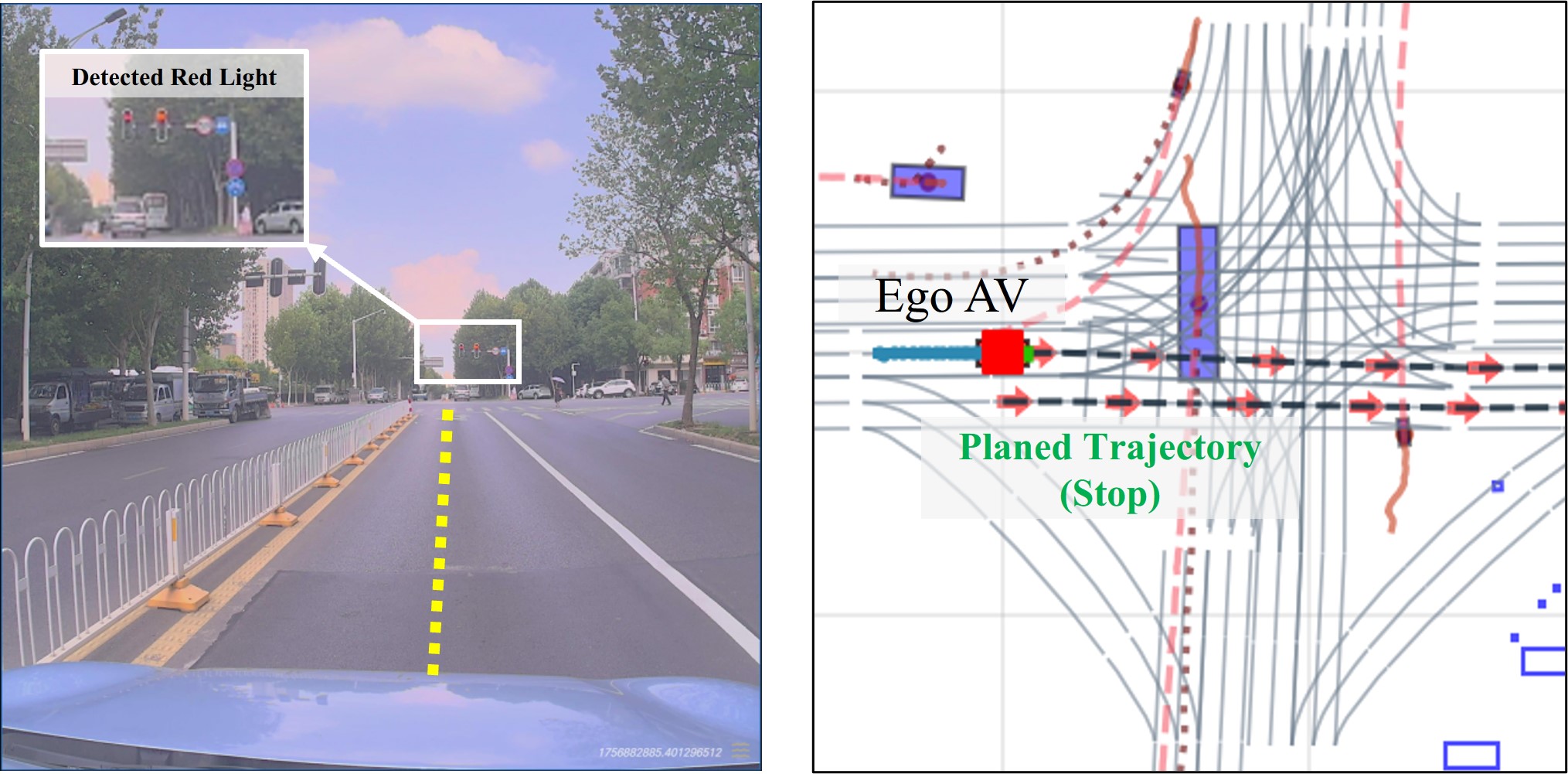}
    \caption{Red-light violation case before and after DTCCL learning. Left: The autonomous bus is passing an intersection with red light detected, the initial policy incorrectly assigns a forward-driving trajectory, leading to disengagement. Right: after DTCCL training, the policy generates a stop trajectory before the intersection, preventing violation.}
    \label{fig:case1}
\end{figure}

\begin{figure}
    \centering
    \includegraphics[width=1\linewidth]{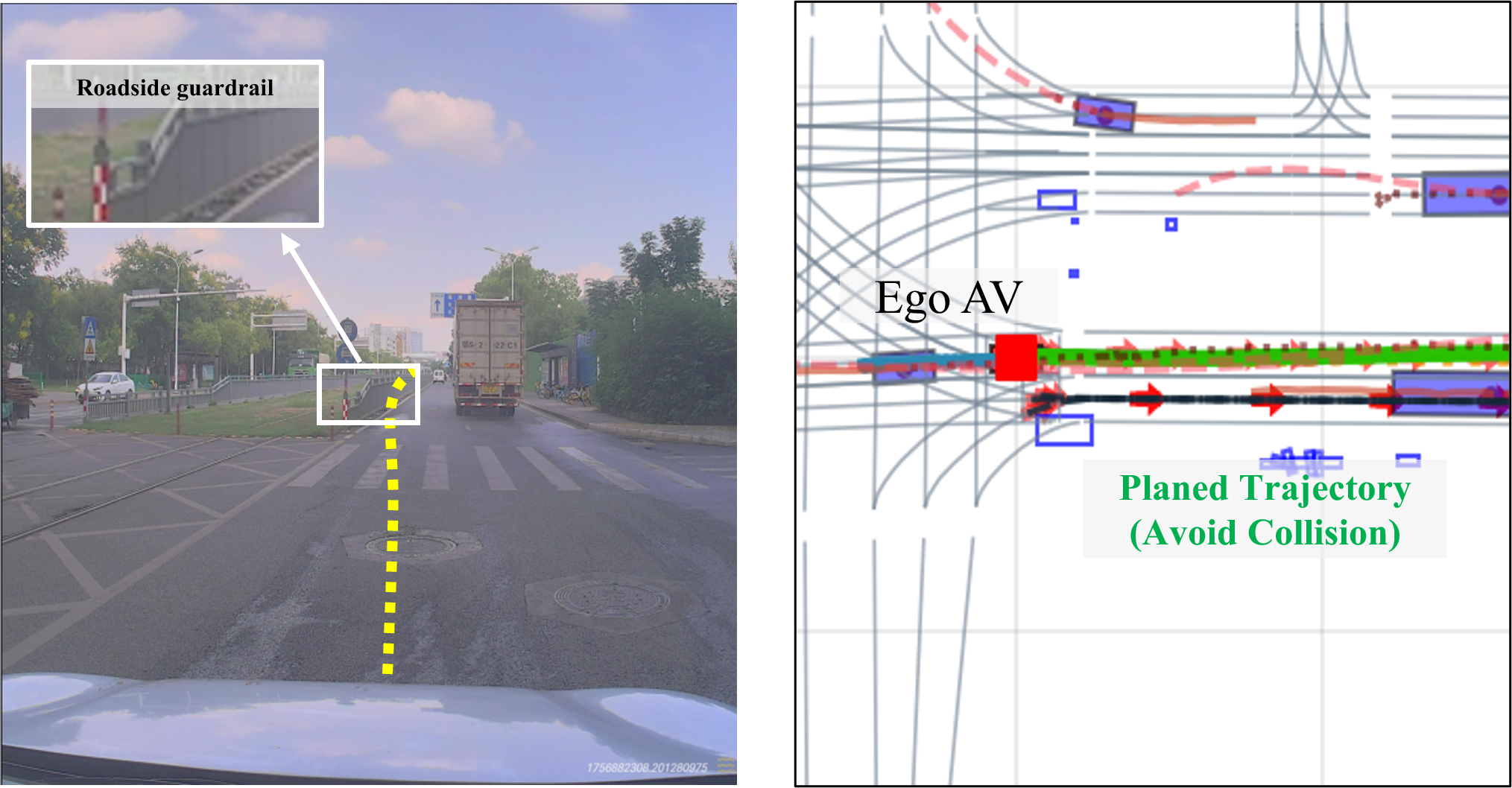}
    \caption{Left: While overtaking a slow-moving vehicle, the initial policy yields an overly conservative leftward deviation, bringing the autonomous bus dangerously close to the roadside guardrail and triggering a disengagement.
Right: After DTCCL learning, the updated policy generates a stable, collision-free trajectory that safely avoids the lead vehicle without encroaching on the guardrail.}
    \label{fig:case2}
\end{figure}



\section{Conclusion and Future Work}

This paper presents a closed-loop continual learning framework for policy improvement in autonomous bus planning. The proposed DTCCL framework leverages real-world disengagement events as high-value feedback to identify policy weaknesses and enable targeted learning. By constructing a “Collect–Augment–Adapt” pipeline, it mines disengagement contexts, generates semantically consistent positive and negative samples through safety-aware perturbations, and integrates contrastive objectives for fine-grained policy refinement. This design effectively mitigates overfitting and enhances robustness in safety-critical long-tail scenarios.

Extensive experiments on the nuPlan dataset and real-world bus disengagement logs demonstrate that DTCCL significantly improves closed-loop driving performance, particularly in challenging urban conditions. The method achieves notable gains in overall safety, stability, and route efficiency, validating the benefits of transforming disengagement data into structured supervision for continuous policy enhancement. 

The proposed approach establishes a foundation for automated cloud–edge optimization in large-scale autonomous bus fleets. In future work, we will explore adaptive weighting strategies for sample importance, extend contrastive learning to multi-agent interaction modeling, and investigate real-time online adaptation mechanisms for broader deployment in complex mixed-traffic environments.



\bibliographystyle{IEEEtran}
\bibliography{Reference}

\begin{thebibliography}{10}
\providecommand{\url}[1]{#1}
\csname url@samestyle\endcsname
\providecommand{\newblock}{\relax}
\providecommand{\bibinfo}[2]{#2}
\providecommand{\BIBentrySTDinterwordspacing}{\spaceskip=0pt\relax}
\providecommand{\BIBentryALTinterwordstretchfactor}{4}
\providecommand{\BIBentryALTinterwordspacing}{\spaceskip=\fontdimen2\font plus
\BIBentryALTinterwordstretchfactor\fontdimen3\font minus \fontdimen4\font\relax}
\providecommand{\BIBforeignlanguage}[2]{{%
\expandafter\ifx\csname l@#1\endcsname\relax
\typeout{** WARNING: IEEEtran.bst: No hyphenation pattern has been}%
\typeout{** loaded for the language `#1'. Using the pattern for}%
\typeout{** the default language instead.}%
\else
\language=\csname l@#1\endcsname
\fi
#2}}
\providecommand{\BIBdecl}{\relax}
\BIBdecl

\bibitem{codevilla2018end}
F.~Codevilla, M.~M{\"u}ller, A.~L{\'o}pez, V.~Koltun, and A.~Dosovitskiy, ``End-to-end driving via conditional imitation learning,'' in \emph{2018 IEEE international conference on robotics and automation (ICRA)}.\hskip 1em plus 0.5em minus 0.4em\relax IEEE, 2018, pp. 4693--4700.

\bibitem{bojarski2016end}
M.~Bojarski, D.~Del~Testa, D.~Dworakowski, B.~Firner, B.~Flepp, P.~Goyal, L.~D. Jackel, M.~Monfort, U.~Muller, J.~Zhang \emph{et~al.}, ``End to end learning for self-driving cars,'' \emph{arXiv preprint arXiv:1604.07316}, 2016.

\bibitem{cao2022autonomous}
Z.~Cao, X.~Li, K.~Jiang, W.~Zhou, X.~Liu, N.~Deng, and D.~Yang, ``Autonomous driving policy continual learning with one-shot disengagement case,'' \emph{IEEE Transactions on Intelligent Vehicles}, vol.~8, no.~2, pp. 1380--1391, 2022.

\bibitem{zhou2025drarl}
W.~Zhou, B.~Zhang, Z.~Cao, X.~Li, Q.~Cheng, C.~Liu, Y.~Zhang, and D.~Yang, ``Drarl: Disengagement-reason-augmented reinforcement learning for efficient improvement of autonomous driving policy,'' \emph{arXiv preprint arXiv:2506.16720}, 2025.

\bibitem{cao2023continuous}
Z.~Cao, K.~Jiang, W.~Zhou, S.~Xu, H.~Peng, and D.~Yang, ``Continuous improvement of self-driving cars using dynamic confidence-aware reinforcement learning,'' \emph{Nature Machine Intelligence}, vol.~5, no.~2, pp. 145--158, 2023.

\bibitem{mei2024continuously}
J.~Mei, Y.~Ma, X.~Yang, L.~Wen, X.~Cai, X.~Li, D.~Fu, B.~Zhang, P.~Cai, M.~Dou \emph{et~al.}, ``Continuously learning, adapting, and improving: A dual-process approach to autonomous driving,'' \emph{arXiv preprint arXiv:2405.15324}, 2024.

\bibitem{liu2021decision}
Q.~Liu, X.~Li, S.~Yuan, and Z.~Li, ``Decision-making technology for autonomous vehicles: Learning-based methods, applications and future outlook,'' in \emph{2021 IEEE International Intelligent Transportation Systems Conference (ITSC)}.\hskip 1em plus 0.5em minus 0.4em\relax IEEE, 2021, pp. 30--37.

\bibitem{kuderer2015learning}
M.~Kuderer, S.~Gulati, and W.~Burgard, ``Learning driving styles for autonomous vehicles from demonstration,'' in \emph{2015 IEEE international conference on robotics and automation (ICRA)}.\hskip 1em plus 0.5em minus 0.4em\relax IEEE, 2015, pp. 2641--2646.

\bibitem{chen2015deepdriving}
C.~Chen, A.~Seff, A.~Kornhauser, and J.~Xiao, ``Deepdriving: Learning affordance for direct perception in autonomous driving,'' in \emph{Proceedings of the IEEE international conference on computer vision}, 2015, pp. 2722--2730.

\bibitem{cheng2024pluto}
J.~Cheng, Y.~Chen, and Q.~Chen, ``Pluto: Pushing the limit of imitation learning-based planning for autonomous driving,'' \emph{arXiv preprint arXiv:2404.14327}, 2024.

\bibitem{cheng2024rethinking}
J.~Cheng, Y.~Chen, X.~Mei, B.~Yang, B.~Li, and M.~Liu, ``Rethinking imitation-based planners for autonomous driving,'' in \emph{2024 IEEE International Conference on Robotics and Automation (ICRA)}.\hskip 1em plus 0.5em minus 0.4em\relax IEEE, 2024, pp. 14\,123--14\,130.

\bibitem{sallab2017deep}
A.~E. Sallab, M.~Abdou, E.~Perot, and S.~Yogamani, ``Deep reinforcement learning framework for autonomous driving,'' \emph{arXiv preprint arXiv:1704.02532}, 2017.

\bibitem{nie2024robust}
Q.~Nie, J.~Ou, H.~Zhang, J.~Lu, S.~Li, and H.~Shi, ``A robust integrated multi-strategy bus control system via deep reinforcement learning,'' \emph{Engineering Applications of Artificial Intelligence}, vol. 133, p. 107986, 2024.

\bibitem{zhou2022dynamically}
W.~Zhou, Z.~Cao, N.~Deng, X.~Liu, K.~Jiang, and D.~Yang, ``Dynamically conservative self-driving planner for long-tail cases,'' \emph{IEEE Transactions on Intelligent Transportation Systems}, vol.~24, no.~3, pp. 3476--3488, 2022.

\bibitem{gu2023safe}
Z.~Gu, L.~Gao, H.~Ma, S.~E. Li, S.~Zheng, W.~Jing, and J.~Chen, ``Safe-state enhancement method for autonomous driving via direct hierarchical reinforcement learning,'' \emph{IEEE Transactions on Intelligent Transportation Systems}, vol.~24, no.~9, pp. 9966--9983, 2023.

\bibitem{naveed2021trajectory}
K.~B. Naveed, Z.~Qiao, and J.~M. Dolan, ``Trajectory planning for autonomous vehicles using hierarchical reinforcement learning,'' in \emph{2021 IEEE International Intelligent Transportation Systems Conference (ITSC)}.\hskip 1em plus 0.5em minus 0.4em\relax IEEE, 2021, pp. 601--606.

\bibitem{parisi2019continual}
G.~I. Parisi, R.~Kemker, J.~L. Part, C.~Kanan, and S.~Wermter, ``Continual lifelong learning with neural networks: A review,'' \emph{Neural networks}, vol. 113, pp. 54--71, 2019.

\bibitem{rusu2016progressive}
A.~A. Rusu, N.~C. Rabinowitz, G.~Desjardins, H.~Soyer, J.~Kirkpatrick, K.~Kavukcuoglu, R.~Pascanu, and R.~Hadsell, ``Progressive neural networks,'' \emph{arXiv preprint arXiv:1606.04671}, 2016.

\bibitem{wu2022prioritized}
J.~Wu, Z.~Huang, W.~Huang, and C.~Lv, ``Prioritized experience-based reinforcement learning with human guidance for autonomous driving,'' \emph{IEEE Transactions on Neural Networks and Learning Systems}, vol.~35, no.~1, pp. 855--869, 2022.

\bibitem{smith2024adaptive}
J.~S. Smith, L.~Valkov, S.~Halbe, V.~Gutta, R.~Feris, Z.~Kira, and L.~Karlinsky, ``Adaptive memory replay for continual learning,'' in \emph{Proceedings of the IEEE/CVF Conference on Computer Vision and Pattern Recognition}, 2024, pp. 3605--3615.

\bibitem{kang2024continual}
D.~Kang, D.~Kum, and S.~Kim, ``Continual learning for motion prediction model via meta-representation learning and optimal memory buffer retention strategy,'' in \emph{Proceedings of the IEEE/CVF Conference on Computer Vision and Pattern Recognition}, 2024, pp. 15\,438--15\,448.

\bibitem{zhang3cil}
H.~Zhang, Z.~Zheng, and X.~Lin, ``3cil: Causality-inspired contrastive conditional imitation learning for autonomous driving.''

\bibitem{ma2022compass}
S.~Ma, S.~Vemprala, W.~Wang, J.~K. Gupta, Y.~Song, D.~McDufft, and A.~Kapoor, ``Compass: Contrastive multimodal pretraining for autonomous systems,'' in \emph{2022 IEEE/RSJ International Conference on Intelligent Robots and Systems (IROS)}.\hskip 1em plus 0.5em minus 0.4em\relax IEEE, 2022, pp. 1000--1007.

\bibitem{xiao2021action}
Y.~Xiao, F.~Codevilla, C.~Pal, and A.~Lopez, ``Action-based representation learning for autonomous driving,'' in \emph{Conference on Robot Learning}.\hskip 1em plus 0.5em minus 0.4em\relax PMLR, 2021, pp. 232--246.

\bibitem{favaro2018autonomous}
F.~Favar{\`o}, S.~Eurich, and N.~Nader, ``Autonomous vehicles’ disengagements: Trends, triggers, and regulatory limitations,'' \emph{Accident Analysis \& Prevention}, vol. 110, pp. 136--148, 2018.

\bibitem{dixit2016autonomous}
V.~V. Dixit, S.~Chand, and D.~J. Nair, ``Autonomous vehicles: disengagements, accidents and reaction times,'' \emph{PLoS one}, vol.~11, no.~12, p. e0168054, 2016.

\bibitem{hu2023simulation}
X.~Hu, S.~Li, T.~Huang, B.~Tang, R.~Huai, and L.~Chen, ``How simulation helps autonomous driving: A survey of sim2real, digital twins, and parallel intelligence,'' \emph{IEEE Transactions on Intelligent Vehicles}, vol.~9, no.~1, pp. 593--612, 2023.

\bibitem{niu2021dr2l}
H.~Niu, J.~Hu, Z.~Cui, and Y.~Zhang, ``Dr2l: Surfacing corner cases to robustify autonomous driving via domain randomization reinforcement learning,'' in \emph{Proceedings of the 5th International Conference on Computer Science and Application Engineering}, 2021, pp. 1--8.

\bibitem{mehta2020active}
B.~Mehta, M.~Diaz, F.~Golemo, C.~J. Pal, and L.~Paull, ``Active domain randomization,'' in \emph{Conference on Robot Learning}.\hskip 1em plus 0.5em minus 0.4em\relax PMLR, 2020, pp. 1162--1176.

\bibitem{zuo2025ralad}
J.~Zuo, H.~Hu, Z.~Zhou, Y.~Cui, Z.~Liu, J.~Wang, N.~Guan, J.~Wang, and C.~J. Xue, ``Ralad: Bridging the real-to-sim domain gap in autonomous driving with retrieval-augmented learning,'' \emph{arXiv preprint arXiv:2501.12296}, 2025.

\bibitem{DAOUD2024184}
\BIBentryALTinterwordspacing
A.~Daoud, C.~Bunel, and M.~Guériau, ``Cornersim: A virtualization framework to generate realistic corner-case scenarios for autonomous driving perception testing,'' \emph{Procedia Computer Science}, vol. 238, pp. 184--191, 2024, the 15th International Conference on Ambient Systems, Networks and Technologies Networks (ANT) / The 7th International Conference on Emerging Data and Industry 4.0 (EDI40), April 23-25, 2024, Hasselt University, Belgium. [Online]. Available: \url{https://www.sciencedirect.com/science/article/pii/S187705092401250X}
\BIBentrySTDinterwordspacing

\bibitem{yu2024using}
B.~Yu, X.~Feng, Y.~Kong, Y.~Chen, Z.~Cheng, and S.~Bao, ``Using meta-learning to establish a highly transferable driving speed prediction model from the visual road environment,'' \emph{Engineering Applications of Artificial Intelligence}, vol. 130, p. 107727, 2024.

\bibitem{10529537}
C.~Chang, S.~Wang, J.~Zhang, J.~Ge, and L.~Li, ``Llmscenario: Large language model driven scenario generation,'' \emph{IEEE Transactions on Systems, Man, and Cybernetics: Systems}, vol.~54, no.~11, pp. 6581--6594, 2024.

\bibitem{karnchanachari2024towards}
N.~Karnchanachari, D.~Geromichalos, K.~S. Tan, N.~Li, C.~Eriksen, S.~Yaghoubi, N.~Mehdipour, G.~Bernasconi, W.~K. Fong, Y.~Guo \emph{et~al.}, ``Towards learning-based planning: The nuplan benchmark for real-world autonomous driving,'' in \emph{2024 IEEE International Conference on Robotics and Automation (ICRA)}.\hskip 1em plus 0.5em minus 0.4em\relax IEEE, 2024, pp. 629--636.

\bibitem{qian2024spider}
Z.~Qian, K.~Jiang, Z.~Cao, K.~Qian, Y.~Xu, W.~Zhou, and D.~Yang, ``Spider: Self-driving planners and intelligent decision-making engines with reusability,'' in \emph{2024 IEEE 27th International Conference on Intelligent Transportation Systems (ITSC)}.\hskip 1em plus 0.5em minus 0.4em\relax IEEE, 2024, pp. 937--944.

\bibitem{caesar2021nuplan}
H.~Caesar, J.~Kabzan, K.~S. Tan, W.~K. Fong, E.~Wolff, A.~Lang, L.~Fletcher, O.~Beijbom, and S.~Omari, ``nuplan: A closed-loop ml-based planning benchmark for autonomous vehicles,'' \emph{arXiv preprint arXiv:2106.11810}, 2021.

\end{thebibliography}

\end{document}